\documentclass[11pt]{article}

\usepackage[preprint]{acl}

\usepackage{times}
\usepackage{latexsym}
\usepackage[T1]{fontenc}
\usepackage[utf8]{inputenc}
\usepackage[protrusion=false]{microtype}
\usepackage{inconsolata}
\usepackage{graphicx}
\usepackage{tikz}
\usepackage{pgfplots}
\pgfplotsset{compat=1.18}
\usepgfplotslibrary{groupplots}
\definecolor{Cbase}{HTML}{BFBFBF}
\definecolor{Cllama}{HTML}{7BAFD4}
\definecolor{Cgemma}{HTML}{82B366}
\definecolor{Cmistral}{HTML}{D4776B}
\definecolor{Czero}{HTML}{A8A8A8}
\definecolor{Cpipe}{HTML}{E8A838}
\pgfplotsset{
  barchartstyle/.style={
    ybar,
    axis on top,
    axis line style={gray!65, line width=0.4pt},
    ymajorgrids=false,
    tick align=outside,
    major tick length=0pt,
    every node near coord/.append style={font=\tiny, /pgf/number format/fixed, /pgf/number format/precision=2},
    tick label style={font=\small},
    label style={font=\small},
    legend cell align=left,
  }
}
\usepackage{booktabs}
\usepackage{multirow}
\usepackage{array}
\usepackage{amsmath}
\usepackage{float}
\DeclareTextFontCommand{\texttt}{\fontencoding{T1}\ttfamily}

\newcolumntype{L}[1]{>{\raggedright\arraybackslash}p{#1}}
\usepackage{listings}
\newcommand{\oursystem}{\textsc{Reap}}

\title{\oursystem{}: Relation-Aware Elicitation and Parsing \\ for Closed-Book Knowledge Base Construction from LLMs}

\author{
  Thanh-Dan Bui \qquad Thanh-Trung Do \qquad Tuan-Phong Nguyen\thanks{\,Corresponding author.} \\
  VNU University of Engineering and Technology, Hanoi, Vietnam \\
  \texttt{\{23020342, 24022472, tuanphong\}@vnu.edu.vn}
}

\begin{document}
\raggedbottom
\maketitle

\begin{abstract}
We present the \oursystem{} system for the AKBC Shared Task 2026 on constructing knowledge bases from language models in a closed-book setting, subject to a budget of at most 32B parameters and no model fine-tuning. Our system combines structured chain-of-thought reasoning, relation-specific query strategies, and a reasoning-based empty-set gate to elicit parametric knowledge, followed by direct extraction into valid JSON arrays. On the official test set, the system, built on Mistral-Small-24B-Instruct-2501, achieves a macro-F1 score of $0.62$, with particularly strong results on \texttt{countryLandBordersCountry} ($F_1=0.95$), \texttt{companyTradesAtStockExchange} ($F_1=0.73$), and \texttt{hasArea} ($F_1=0.77$). Our code is publicly available.\footnote{ \url{https://github.com/yammdd/AKBC-Shared-Task-2026}}

\end{abstract}

\section{Introduction}
\label{sec:intro}

Large language models (LLMs) encode substantial factual knowledge in their parameters \citep{petroni2019lama}. However, eliciting this knowledge for knowledge base construction (KBC) is considerably more challenging than answering isolated factual questions: a subject--relation pair $(s,r)$ may correspond to an empty ($\emptyset$), single-valued ($1$), or multi-valued ($N$) object set. The AKBC Shared Task 2026 \citep{lmkbc2026} formalizes this problem under a constrained setting: systems must operate in a closed-book fashion, without retrieval-augmented generation or external knowledge; use at most 32B parameters; and perform no model fine-tuning.

In this task, each input record specifies a subject entity and one of the six given relations (see Table~\ref{tab:reldef}), and the system must return the complete set of corresponding objects as a valid JSON array. The relations cover geographic adjacency and area, a person's city of death, venue capacity, award recipients, and the stock exchanges on which a company is traded. Consequently, the required output may be empty, contain a single object, or contain a long list of objects, while quantitative answers must also be sufficiently precise.
Table~\ref{tab:reldef} summarizes the six target relations and the scope of their ground-truth labels.

\begin{table*}[t]
\centering
\captionsetup{justification=centering}
\small
\setlength{\tabcolsep}{6pt}
\renewcommand{\arraystretch}{1.1}
\begin{tabular}{@{}L{5.0cm}L{10.5cm}@{}}
\toprule
\textbf{Relation} & \textbf{Ground-truth scope and definition} \\
\midrule
\texttt{countryLandBordersCountry} & Countries that share a land border with the subject (maritime borders excluded; island countries $\to \emptyset$). \\
\texttt{personHasCityOfDeath} & The city in which the person died (living people or cases with an unknown city $\to \emptyset$). \\
\texttt{hasCapacity} & The maximum spectator capacity of the venue, represented as an integer. \\
\texttt{awardWonBy} & Entities, including individuals and organizations, that have received the corresponding award. \\
\texttt{companyTradesAtStockExchange} & Stock exchanges on which the company's shares are publicly traded (private companies $\to \emptyset$). \\
\texttt{hasArea} & The total surface area (land $+$ inland water) of the geographical entity, measured in km\textsuperscript{2}. \\
\bottomrule
\end{tabular}
\caption{Definitions and ground-truth scope of the six target relations.}
\label{tab:reldef}
\end{table*}

Earlier factual probing methods use cloze-style prompts or prompt ensembles to rank candidate objects \citep{petroni2019lama,jiang2020lpaqa}, while the LM-KBC challenge series additionally requires systems to materialize variable-cardinality answer sets \citep{lmkbc2022,lmkbc2023,lmkbc2024,lmkbc2025}. A direct generative solution that asks the model to recall all objects and produce correctly formatted JSON in a single pass can therefore hallucinate
objects, omit valid answers, or violate the output schema. 

To address this issue, we propose \oursystem{} (\textbf{R}elation-aware \textbf{E}licitation \textbf{A}nd \textbf{P}arsing), a two-stage pipeline that decouples factual elicitation from answer serialization. Stage~1 prompts the LLM using strategies tailored to each relation, whereas Stage~2 uses deterministic JSON parsing when possible and LLM-based extraction only as a fallback. This separation allows the model to focus on recalling the answer set before the system enforces the required JSON format.

Our main contributions are as follows:
\begin{enumerate}
    \item \textbf{Relation-Aware Elicitation}: We design specialized strategies for different relation types, including CoT reasoning with an empty-set gate, a four-direction geographical scan for land borders, and chronological multi-pass querying for awards.
    \item \textbf{Efficient Hybrid Parsing}: Deterministic parsing handles approximately $80\%$ of the records, while LLM-based extraction is reserved for complex outputs, substantially reducing the overall computational cost.
    \item \textbf{Scalable Inference and Evaluation}: Batched inference with vLLM on TPU hardware enables fast end-to-end processing. The complete system achieves macro-F1 $0.62$ on the test set and highlights the limits of parametric knowledge for rare entities.
\end{enumerate}

\section{Related Work}
\label{sec:related}

\paragraph{Probing knowledge in language models.}
LAMA \citep{petroni2019lama} established cloze-style probing but assumes exactly one object per relation, and what can be extracted depends heavily on the prompt \citep{jiang2020lpaqa}; see \citet{alkhamissi2022review} for a survey. The LM-KBC challenge series \citep{lmkbc2022, lmkbc2023, lmkbc2024, lmkbc2025} recast this as knowledge base construction with arbitrary-cardinality object sets ($0$, $1$, or $N$), the setting continued by the AKBC Shared Task 2026 \citep{lmkbc2026}. Prior systems pair LLM probing with entity mapping \citep{zhang2023llmke}; most recently, ReWiSe \citep{rewise2025}, a top-performing system at LM-KBC 2025, combines chain-of-thought (CoT) reasoning with relation-wise self-consistency. In \oursystem{}, we adopt the idea of relation-specific CoT and explicit empty-set handling, but drop self-consistency voting, which is costly and can be dominated by confident-but-incorrect reasoning chains.

\paragraph{Parametric knowledge and its limits.}
Closed-book QA shows that language models can answer factual questions without retrieval \citep{roberts2020closedbook}, motivating the closed-book setting we adopt. This parametric knowledge is uneven, however: models recall popular facts well but struggle with long-tail entities \citep{kandpal2023longtail, mallen2023trust}, which bounds any closed-book system and matches our error analysis on rare subjects.

\paragraph{Step-by-step reasoning and empty-set gate.}
Chain-of-thought prompting \citep{wei2022cot, kojima2022zeroshot} improves multi-step reasoning and builds on the in-context learning ability of instruction-tuned LLMs \citep{brown2020gpt3}. 
Since a correct answer may itself be the empty set, in \oursystem{}, we add a \emph{reasoning-based empty-set gate} inspired by selective prediction \citep{kamath2020selective}. Unlike calibrated methods that estimate whether a model knows an answer \citep{kadavath2022know}, our gate is a prompt-driven decision that returns an empty set as the final answer when the property does not exist or the evidence of it is insufficient.

\section{Method}
\label{sec:method}

\subsection{Architecture Overview}

\begin{figure}[t]
\centering
\includegraphics[width=0.95\columnwidth]{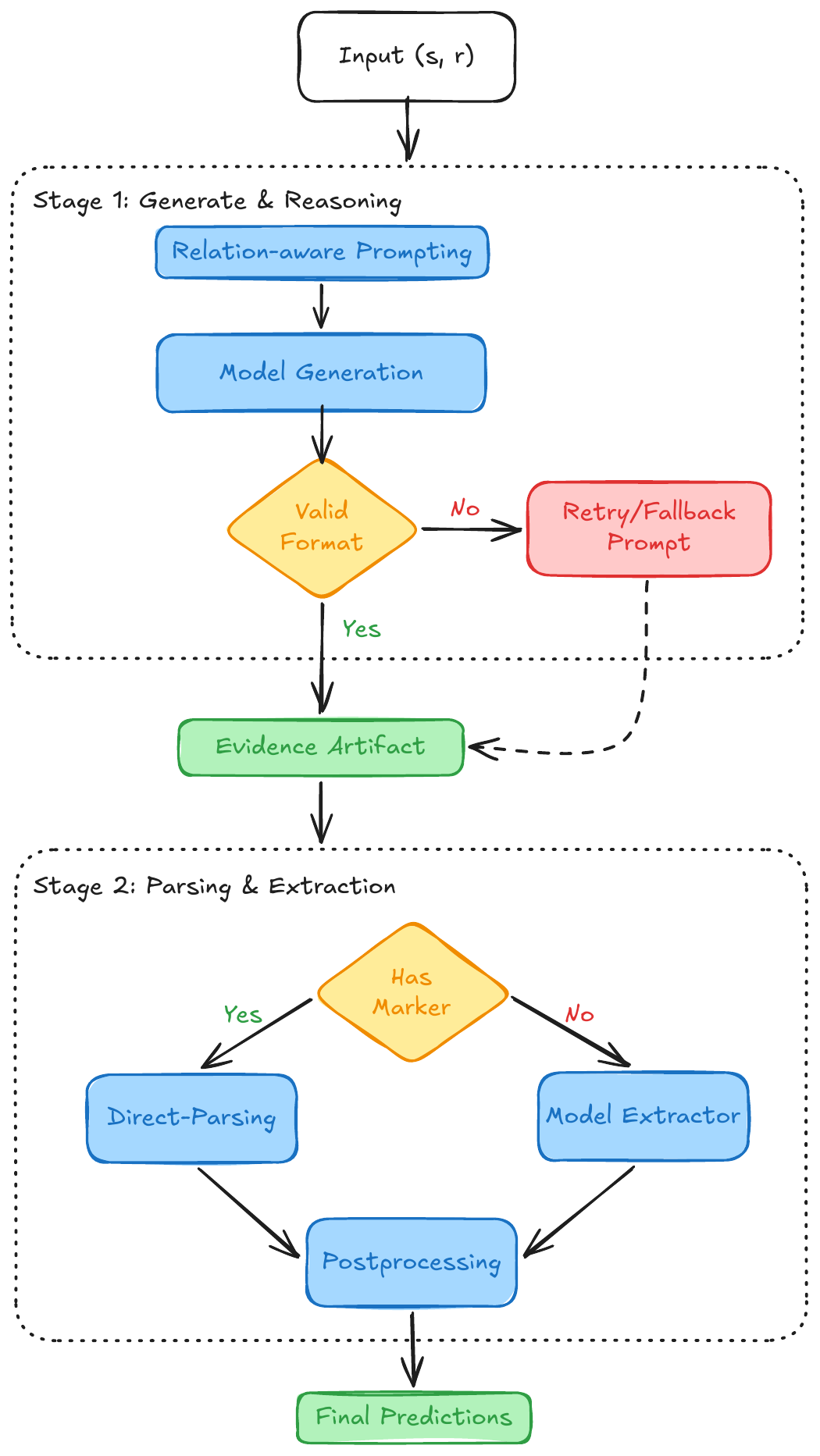}
\caption{The architecture of \oursystem{}.}
\label{fig:arch}
\end{figure}

The \oursystem{} system consists of two sequential stages (Figure~\ref{fig:arch}). In Stage~1, we use an LLM to \emph{generate evidence} as free-form text for multi-valued relations or use CoT reasoning for single-valued relations. If the generation phase produces unparsable output, the system automatically retries or switches to a fallback prompt. Stage~2 \emph{extracts} the evidence into JSON arrays using regular expressions; for multi-valued relations that cannot be handled reliably by simple rules, the LLM performs the extraction itself.

\subsection{Relation-Specific Prompt Design}
\label{sec:promptdesign}

Stage~1 prompts are specialized per relation following three principles: 
(i) \textbf{query decomposition} to maximize recall on multi-valued relations (e.g., six decade-scoped passes for \texttt{awardWonBy} and a four-direction geographical scan for \texttt{countryLandBordersCountry}); 
(ii) \textbf{output-format constraints} via \texttt{BORDERS:} or \texttt{FINAL\_ANSWER: [\ldots]} markers; and 
(iii) \textbf{step-by-step reasoning with an empty-set gate} that outputs \texttt{FINAL\_ANSWER: []} for unanswerable entities (e.g., living individuals or private companies). Together, these relation-tailored, multi-step CoT procedures constitute the \emph{structured reasoning} that Stage~1 performs before extraction.

Few-shot demonstrations are static exemplar pairs (one positive, one empty-set case) fixed per relation to illustrate the reasoning chain and empty-set triggering. Crucially, all exemplars are drawn exclusively from \texttt{train.jsonl} or hand-written triples, ensuring zero data leakage against the evaluation splits.

\paragraph{\texttt{awardWonBy}: Chronological Multi-Pass.}
For major awards with dozens or hundreds of recipients, a single query typically yields low recall. We therefore use a multi-pass strategy comprising one general prompt and five auxiliary prompts decomposed chronologically: from the award's inception through the 1970s, the 1980s--1990s, the 2000s--2010s, 2020 onward, and a final pass targeting less prominent or non-Western recipients. All six generations use temperature $T=0.7$ to diversify recall before the evidence is merged.

\paragraph{\texttt{countryLandBordersCountry}: Four-Direction Geographical Scan.}
The prompt instructs the model to act as a geography expert and inspect the subject's eastern, western, southern, and northern boundaries. This scan reduces omissions of small neighboring countries and adjacent detached territories. The prompt includes land borders only, explicitly excludes maritime boundaries, and returns \texttt{BORDERS: NONE} for island countries. If the response lacks the \texttt{BORDERS:} marker, the system automatically retries with a shorter fallback prompt.

\paragraph{\texttt{companyTradesAtStockExchange}: Four-Step Listing Check.}
The model follows a four-step CoT procedure: (1) identify the company; (2) apply a \textbf{Strict Public Check (empty-set gate)} to determine whether the entity is a private company, a nonprofit organization, or a subsidiary that is not separately listed, returning \texttt{FINAL\_ANSWER: []} if any condition holds; (3) retrieve all listing exchanges using their full official names, such as \textit{New York Stock Exchange} or \textit{SIX Swiss Exchange}; and (4) conclude with \texttt{FINAL\_ANSWER: ["Exchange Name"]}.

\paragraph{\texttt{hasArea}: Four-Step Entity Disambiguation and Unit Normalization.}
The model follows four CoT steps: (1) distinguish the target entity (e.g., an island, lake, or region) from the country containing it; (2) retrieve its area from parametric knowledge while preserving decimal precision; (3) convert the value to $\text{km}^2$; and (4) output an integer or decimal value as \texttt{FINAL\_ANSWER: ["Value"]}.

\paragraph{\texttt{hasCapacity}: Four-Step Reasoning with Capacity-Range Verification.}
The model follows four CoT steps: (1) determine the venue type and location; (2) distinguish a university or small local stadium from a larger national stadium with a similar name or in the same city; (3) check the estimate against a plausible capacity range ($1{,}000$--$35{,}000$ seats for small venues and $35{,}000$--$100{,}000$ for large national stadiums); and (4) return the capacity as an integer in \texttt{FINAL\_ANSWER: ["Value"]}.

\paragraph{\texttt{personHasCityOfDeath}: Five-Step Reasoning with Empty-Set Gate.}
The model performs five steps: (1) check the subject's biographical status and record ``ALIVE TODAY'' if appropriate; (2) resolve name ambiguity; (3) retrieve the city or town of death; (4) isolate the place name; and (5) return \texttt{FINAL\_ANSWER: []} if the person is alive or the information is unavailable, or \texttt{FINAL\_ANSWER: ["CityName"]} if the place of death is confirmed.

\subsection{Post-processing}
\label{sec:postprocessing}

To ensure that extracted outputs adhere to the ground-truth format, we apply the following automated post-processing procedure after the generation phase:
\begin{enumerate}
    \item \textbf{Direct Parsing and Robust Array Extraction}: The system directly parses the content of the \texttt{FINAL\_ANSWER:} line or recovers a truncated JSON array using balanced-bracket scanning.
    \item \textbf{Numeric Extraction and Normalization}: For quantitative relations (\texttt{hasArea} and \texttt{hasCapacity}), regular expressions extract integer or floating-point values while removing thousands separators and measurement units.
    \item \textbf{Title and Noise Filtering}: For \texttt{awardWonBy}, the system removes year prefixes (e.g., \textit{"1938: Albert Einstein"}), HTML tags, work-title suffixes, and honorifics such as \textit{Dr.}, \textit{Prof.}, \textit{Sir}, \textit{Dame}, \textit{Lord}, \textit{Saint}, \textit{St.}, and \textit{Mr.} to obtain normalized entity names.
    \item \textbf{Parenthetical Filtering and Deduplication}: The system removes trailing parenthetical qualifiers (e.g., \textit{"Guinea (West Africa)"} $\to$ \textit{"Guinea"}) and performs case-insensitive deduplication while preserving the capitalization of the first occurrence.
\end{enumerate}

\section{Experiments}
\label{sec:experiment}

\subsection{Experimental Setup}


\paragraph{Data.}
The organizers provide training, validation, and test splits for six relations. The validation and test splits each contain 475 records: 68 for \texttt{countryLandBordersCountry} in validation (67 in test), 97 and 98 for \texttt{hasCapacity} respectively, 10 for \texttt{awardWonBy}, and 100 for each of the other three relations. Each record consists of a subject, a relation, and a set of objects, each accompanied by a list of aliases. The data cover all three cardinality cases: empty sets (e.g., \textit{New Zealand} for land borders), single-valued relations, and multi-valued relations (e.g., Nobel Prize in Physics, $229$ recipients in \texttt{train.jsonl}, or AAAI Fellow, 350 people in \texttt{val.jsonl}).

\paragraph{Models.}
We evaluate three instruction-tuned models---Gemma-2-9B-it \citep{gemma2}, Llama-3.1-8B-Instruct \citep{llama3}, and Mistral-Small-24B-Instruct-2501 \citep{mistralsmall24b}---and compare against the organizer's Qwen3.5-9B baseline \citep{qwen3.5}. 
Gemma-2-9B-it and Llama-3.1-8B-Instruct serve as comparison systems, while Mistral-Small-24B-Instruct-2501 (hereafter Mistral-24B) is the primary model in our final system.

\paragraph{Hardware and Runtime.}
Experiments are conducted on Kaggle using a TPU v5e-8 with eight TPU v5e cores. We serve the models with vLLM TPU Server and use bfloat16 precision. With direct parsing and batched inference, one complete run over all six relations in the validation set (475 records) takes approximately \emph{2--5 minutes} with \texttt{batch\_size=32}.

\paragraph{Metrics.}
We report macro/micro precision, recall, and F1 using the organizer-provided \texttt{evaluate.py} script. For string-valued relations, the evaluator normalizes predictions and matches aliases through maximum bipartite matching. For quantitative relations, it applies a relative tolerance of $5\%$. Overall macro-F1 is the mean of the \emph{per-record} F1 scores.

\subsection{Validation Results}
\label{sec:pipeline_results}

\begin{figure}[t]
\centering
\makebox[\columnwidth][c]{%
\begin{tikzpicture}
\begin{axis}[
    barchartstyle,
    width=1.05\columnwidth,
    height=6.2cm,
    bar width=11pt,
    enlarge x limits=0.2,
    ylabel={Macro-F1},
    symbolic x coords={Llama-3.1-8B, Gemma-2-9B, Mistral-24B},
    xtick=data,
    x tick label style={rotate=15, anchor=north east, font=\scriptsize},
    ymin=0, ymax=0.72,
    ytick={0,0.1,...,0.7},
    nodes near coords={\pgfmathprintnumber[fixed,fixed zerofill,precision=2]{\pgfplotspointmeta}},
    every node near coord/.append style={font=\fontsize{5}{5}\selectfont, anchor=south, yshift=1pt},
    legend style={at={(0.5,1.02)}, anchor=south, legend columns=2,
                  font=\scriptsize, draw=none, fill=none,
                  /tikz/every even column/.append style={column sep=8pt}},
]
\addplot[fill=Czero, draw=gray!75, line width=0.4pt, bar shift=-6.5pt] coordinates
    {(Llama-3.1-8B,0.376) (Gemma-2-9B,0.420) (Mistral-24B,0.432)};
\addplot[fill=Cpipe, draw=Cpipe!65!black, line width=0.4pt, bar shift=6.5pt] coordinates
    {(Llama-3.1-8B,0.482) (Gemma-2-9B,0.508) (Mistral-24B,0.645)};
\legend{Zero-shot, \oursystem{} pipeline (ours)}
\end{axis}
\end{tikzpicture}%
}
\caption{Overall macro-F1 of zero-shot prompting and the two-stage pipeline on the validation set.}
\label{fig:zeroshot_vs_pipeline}
\end{figure}
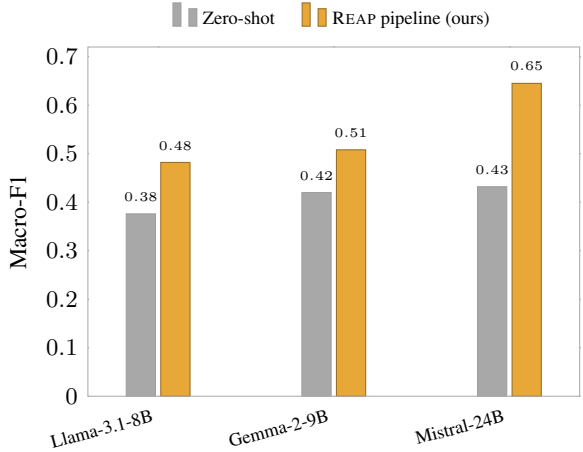

We first evaluate all models in a \emph{zero-shot} setting, using direct queries without evidence generation or CoT reasoning. As shown in Figure~\ref{fig:zeroshot_vs_pipeline}, Llama-3.1-8B, Gemma-2-9B, and Mistral-24B achieve macro-F1 scores of $0.38$, $0.42$, and $0.43$, respectively. These relatively low and similar scores suggest that all three models struggle when required to recall facts and satisfy the output format simultaneously. The limitation is particularly pronounced for multi-valued relations that require long object lists, such as \texttt{awardWonBy}, and for quantitative relations requiring precise values.

\paragraph{Effectiveness of the Two-Stage Pipeline.}
Figure~\ref{fig:zeroshot_vs_pipeline} shows that the two-stage pipeline improves macro-F1 for all three models. Llama-3.1-8B improves from $0.38$ to $0.48$, Gemma-2-9B from $0.42$ to $0.51$, and Mistral-24B from $0.43$ to $0.65$. Mistral-24B yields the largest absolute gain ($+0.22$) and is therefore selected as the primary model in our final system submitted to the AKBC Shared Task 2026.

\subsection{Test Results}

Table~\ref{tab:test_results} reports the relation-level and overall macro-F1 scores of the primary system on the official test set.

\begin{table}[t]
  \centering
  \footnotesize
  \caption{Macro-F1 of Mistral-24B on the test set.}
  \label{tab:test_results}
  \resizebox{\columnwidth}{!}{%
    \begin{tabular}{@{}lc@{}}
      \toprule
      \textbf{Relation} & \textbf{Macro-F1} \\
      \midrule
      \texttt{awardWonBy} & 0.37 \\
      \texttt{companyTradesAtStockExchange} & 0.73 \\
      \texttt{countryLandBordersCountry} & 0.95 \\
      \texttt{hasArea} & 0.77 \\
      \texttt{hasCapacity} & 0.23 \\
      \texttt{personHasCityOfDeath} & 0.53 \\
      \midrule
      \textbf{Overall} & \textbf{0.62} \\
      \bottomrule
    \end{tabular}%
  }
\end{table}

\begin{figure*}[t]
\centering
\begin{tikzpicture}
\begin{axis}[
    barchartstyle,
    width=0.98\textwidth,
    height=7.0cm,
    bar width=7.5pt,
    enlarge x limits=0.065,
    ylabel={Macro-F1},
    symbolic x coords={
      awardWonBy, companyTrades, countryBorders,
      hasArea, hasCapacity, cityOfDeath, Overall},
    xtick=data,
    x tick label style={rotate=18, anchor=north east, font=\footnotesize},
    ymin=0, ymax=1.0,
    ytick={0,0.2,...,1.0},
    legend style={at={(0.5,1.02)}, anchor=south, legend columns=2,
                  font=\scriptsize, draw=gray!45, fill=white,
                  rounded corners=1pt, inner sep=3pt,
                  /tikz/every even column/.append style={column sep=10pt}},
]
\addplot[fill=Cbase, draw=gray!70, line width=0.35pt] coordinates
    {(awardWonBy,0.1510) (companyTrades,0.2727) (countryBorders,0.7315)
     (hasArea,0.3400) (hasCapacity,0.1224) (cityOfDeath,0.1700) (Overall,0.2964)};
\addplot[fill=Cllama, draw=Cllama!65!black, line width=0.35pt] coordinates
    {(awardWonBy,0.3180) (companyTrades,0.5908) (countryBorders,0.7895)
     (hasArea,0.4800) (hasCapacity,0.1837) (cityOfDeath,0.4800) (Overall,0.4824)};
\addplot[fill=Cgemma, draw=Cgemma!65!black, line width=0.35pt] coordinates
    {(awardWonBy,0.2683) (companyTrades,0.6293) (countryBorders,0.9138)
     (hasArea,0.4900) (hasCapacity,0.1429) (cityOfDeath,0.4000) (Overall,0.4839)};
\addplot[
    fill=Cmistral,
    draw=Cmistral!65!black,
    line width=0.35pt,
    nodes near coords={\pgfmathprintnumber[fixed,fixed zerofill,precision=2]{\pgfplotspointmeta}},
    every node near coord/.append style={font=\tiny, anchor=south, yshift=1pt},
] coordinates
    {(awardWonBy,0.3691) (companyTrades,0.7343) (countryBorders,0.9463)
     (hasArea,0.7700) (hasCapacity,0.2347) (cityOfDeath,0.5300) (Overall,0.6179)};
\legend{Organizer's baseline (Qwen3.5-9B), Llama-3.1-8B, Gemma-2-9B, Mistral-24B (ours)}
\draw[dashed, gray!55, line width=0.45pt]
      ([xshift=27pt]axis cs:cityOfDeath,0)
      -- ([xshift=27pt]axis cs:cityOfDeath,1.0);
\end{axis}
\end{tikzpicture}
\caption{Relation-level macro-F1 of the organizer's Qwen3.5-9B baseline and three models using the two-stage \oursystem{} pipeline on the official test set of the AKBC Shared Task 2026.}
\label{fig:pipeline_perf}
\end{figure*}
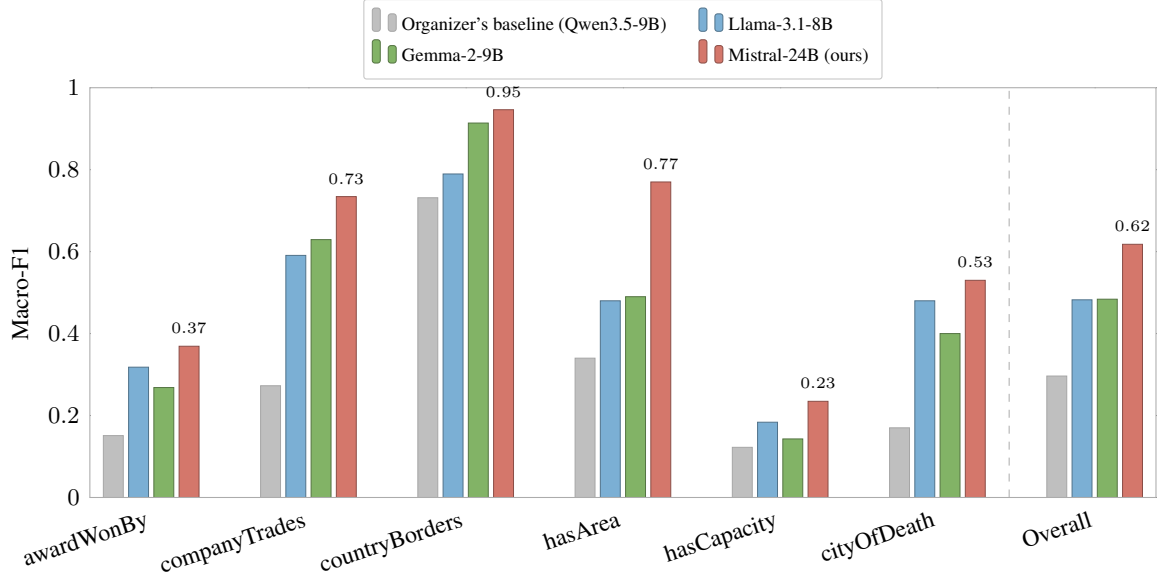

The relation-level results in Figure~\ref{fig:pipeline_perf} show that the Mistral-24B pipeline consistently outperforms all other configurations across all six relations on the F1 metric. It achieves its highest scores on \texttt{countryLandBordersCountry} ($0.95$) and \texttt{hasArea} ($0.77$). Overall, the Mistral pipeline leads with a macro-F1 score of $0.62$, significantly outperforming the Llama ($0.48$) and Gemma ($0.48$) pipelines, as well as more than doubling the organizer's baseline that is based on Qwen3.5-9B ($0.30$).

\section{Ablation Study}
\label{sec:ablation}

To quantify each component's contribution, we perform a subtractive ablation study on the validation set using Mistral-24B (Table~\ref{tab:ablation}).

\begin{table}[t]
  \centering
  \footnotesize
  \caption{Ablation results on the validation set.}
  \label{tab:ablation}
  \resizebox{\columnwidth}{!}{%
    \begin{tabular}{@{}lcc@{}}
      \toprule
      \textbf{Configuration} & \textbf{Macro-F1} & $\mathbf{\Delta}$ \\
      \midrule
      \textbf{Full Pipeline (Ours)} & \textbf{0.65} & -- \\
      \quad w/o Task Decomposition & 0.59 & $-0.06$ \\
      \quad w/o Empty-Set Gate & 0.57 & $-0.08$ \\
      \quad w/o CoT (Zero-Shot) & 0.43 & $-0.22$ \\
      \bottomrule
    \end{tabular}%
  }
\end{table}

Removing \emph{Task Decomposition} drops macro-F1 from $0.65$ to $0.59$, as single-pass queries suffer severe recall loss on multi-valued relations: \texttt{awardWonBy} collapses from $0.25$ to $0.00$ and \texttt{countryLandBordersCountry} falls from $0.98$ to $0.63$. 
Removing the \emph{Empty-Set Gate} lowers macro-F1 to $0.57$ due to hallucinated answers for unanswerable entities, severely hurting precision on \texttt{personHasCityOfDeath} ($0.50 \to 0.36$) and \texttt{companyTradesAtStockExchange} ($0.76 \to 0.57$). 
Finally, removing \emph{CoT Reasoning} entirely causes the largest drop to $0.43$, heavily degrading \texttt{hasArea} ($0.82 \to 0.55$) and \texttt{countryLandBordersCountry} ($0.98 \to 0.27$), confirming that our gains stem primarily from structured parametric reasoning rather than output parsing alone.

\section{Discussion}
\label{sec:discussion}

\paragraph{Relation-Aware Prompting versus Zero-Shot Prompting.}
In the zero-shot setting, all models exhibit limited and relatively similar performance. Requiring a model to recall facts and produce valid JSON in a single pass is especially difficult for relations with large object sets, such as \texttt{awardWonBy}, and for quantitative relations that demand precise values, such as \texttt{hasArea} and \texttt{hasCapacity}. Applying the complete \emph{two-stage pipeline} of \oursystem{} with relation-specific prompts, CoT reasoning, a reasoning-based empty-set gate, and direct parsing improves all three models. These results indicate that structured reasoning is important for reliably eliciting knowledge encoded in model parameters.

\paragraph{Limits of Parametric Knowledge.}
Our results reveal a clear gap between the 9B model (Gemma-2-9B) and the 24B model (Mistral-24B). In the zero-shot setting, the models perform similarly, partly because both struggle with output formatting. Under the full structured-reasoning pipeline, however, Mistral-24B reaches macro-F1 $0.65$ on the validation set, compared with $0.51$ for Gemma-2-9B. This result suggests that reasoning prompts can reduce formatting noise, while successful factual retrieval still depends on model capacity. Mistral-24B performs better on long-tail entities such as small islands, club-level stadiums, and less prominent awards. Thus, in a closed-book setting, a larger model may benefit more from such a pipeline.

\paragraph{Run-to-run variance.}
Two minor sources of variation affect our runs. First, \texttt{awardWonBy} uses sampling ($T=0.7$, six decade-scoped passes) to diversify recall over long recipient lists, so its score fluctuates slightly given only 10 validation awards. Second, even under greedy decoding ($T=0.0$) the TPU v5e-8 is not bit-exact, because the accumulation order of bfloat16 values differs across its eight cores. Both effects are small---overall macro-F1 varies by about $\pm 0.005$---and do not change our conclusions.

\section{Conclusion}
\label{sec:conclusion}

We presented the \oursystem{} system for the AKBC Shared Task 2026 under a strict closed-book setting with at most 32B parameters, no external knowledge, and no model fine-tuning. The system implements a unified two-stage pipeline with Mistral-Small-24B-Instruct-2501, combining relation-aware prompt engineering, a reasoning-based empty-set gate, and direct parsing. It achieves a macro-F1 of $0.65$ on the validation set and a macro-F1 of $0.62$ on the official test set. The results demonstrate that structured reasoning can effectively elicit factual knowledge encoded in model parameters for knowledge base construction.

\section*{Limitations}

Despite its promising performance, our system has several limitations: (i) its parametric knowledge remains incomplete for very rare, long-tail entities; (ii) predictions for quantitative relations sometimes fall outside the $5\%$ relative-tolerance threshold because of retrieval or rounding errors; and (iii) stochastic sampling for \texttt{awardWonBy}, together with non-deterministic distributed computation on TPU v5e-8, causes slight variation across runs. Among the six relations, \texttt{hasCapacity} is the most challenging (test macro-F1 $0.23$): venues are frequently confused with larger namesakes in the same city, and even small numeric errors exceed the $5\%$ tolerance.

\bibliography{custom}

\end{document}